\documentclass{article}

\usepackage{arxiv}

\usepackage[utf8]{inputenc} 
\usepackage[T1]{fontenc}    
\usepackage{hyperref}       
\usepackage{url}            
\usepackage{booktabs}       
\usepackage{amsfonts}       
\usepackage{nicefrac}       
\usepackage{microtype}      
\usepackage{lipsum}
\usepackage{graphicx}
\graphicspath{ {./images/} }

\title{Predicting COVID-19 and pneumonia complications from admission texts}

\author{
 Dmitriy Umerenkov \\
  Sberbank AI Lab\\
  \texttt{d.umerenkov@gmail.com} \\
   \And
 Oleg Cherkashin \\
  Krasnoyarsk Regional Clinical Hospital\\
  \texttt{oacherkashin@gmail.com} \\
  \And
 Alexander Nesterov \\
  Sberbank AI Lab\\
  \texttt{ainesterov@sberbank.ru} \\
   \And
 Victor Gombolevskiy \\
 Artificial Intelligence Research Institute \\
  \texttt{gombolevskiy@airi.net} \\
   \And
 Irina Demko \\
  Krasnoyarsk Regional Clinical Hospital\\
  \texttt{demko64@mail.ru} \\
  \And
 Alexander Yalunin \\
  Sberbank AI Lab\\
  \texttt{ale.yalunin@gmail.com} \\
  \And
 Vladimir Kokh \\
  Sberbank AI Lab \\
  \texttt{kokh.v.n@sberbank.ru} \\
}

\begin{document}
\maketitle
\begin{abstract}
In this paper we present a novel approach to risk assessment for patients hospitalized with pneumonia or COVID-19 based on their admission reports. We applied a Longformer neural network to admission reports and other textual data available shortly after admission to compute risk scores for the patients. 
We used patient data of multiple European hospitals to demonstrate that our approach outperforms the Transformer baselines. Our experiments show that the proposed model generalises across institutions and diagnoses.  
Also, our method has several other advantages described in the paper ----------
\end{abstract}

\maketitle

\section{Introduction}
Risk assessment for admitted patients is an important task especially so in times of pandemic when medical institutions and specialists are severely overburdened. Prioritising high-risk patients and deprioritising low-risk ones is of paramount importance to ensure the maximum effectiveness of limited medical resources.
Admission reports contain a plethora of information. Combined with radiology reports and laboratory results received during the first 24 hours from admission this information is sufficient for experienced medical professionals to estimate the patient severity.
Unfortunately at the peak of pandemic doctors not specialized in respiratory diseases are often forced to make such decisions which can lead to sub-optimal risk assessment and patient routing. 

In this paper we propose a novel method to predict the risk of death or the need for artificial lung ventilation (ALV) for patients hospitalized with pneumonia or COVID-19 based on their admission reports and possibly other information available in first 24 hours of hospital stay. 
We do it by applying a Longformer model with a combination of local and global attention, as opposed to more traditional deep learning methods (convolutional neural networks (CNN), recurrent neural networks (RNN) or Transformer models with global attention only).

Our proposed approach has the following advantages in comparison with current deep learning - based risk assessment methods. 
First as it is shown in the paper the Longformer architecture outperforms the baselines including BERT and a combination of BERT and recurrent neural networks.
Second the proposed model works directly on texts as imported from hospital Electronic Health Records (EHR) systems and do not require any feature engineering or additional clinical features (age,temperature, blood pressure, etc.) as input. This allows for very quick and easy integration of our model into existing medical workflows.
Third the proposed model is robust to changes between institutions, so a model trained on data from several major hospitals can be quickly deployed to other hospitals without the need to retrain the model.
Fourth the model generalises between diseases, so a model trained on admission reports of one disease (in this work pneumonia) can be used to access the risks of other diseases (in this work COVID-19) with somewhat reduced efficiency.
Finally our model has been validated in a hospital setting, with doctors confirming its usefulness in selecting high-risk cases which are sometimes not obvious at admission even to experienced medical professionals.

One issue with the proposed approach is the lack of interpretability of the black-box models such as neural networks. In normal times such lack of interpretability is commonly considered a "no-go" for medical models but in times of pandemic and severe shortages of medical personnel and resources the benefits of our approach can outweigh the risks. 

This paper makes the following contributions. First, we propose to use pretrained Longformer models on long medical texts, applying the models to patient admission reports and other textual data. Second we use raw text data without any any additional clinical features for risk assessment of pneumonia and COVID-19 patients. Third, we tested our method against the benchmarks on the historical data and achieved superior performance. 

The rest of the paper is organised as follows. In Section 2 we discuss the related work. In Section 3 we describe the data used and in Section 4 our proposed method. In Section 5 we present our experiments and Section 6 is dedicated to the discussion of our results and conclusions.

\section{Related work}

Clinical decision support (CDS) systems are tools providing assistance in decision making to address challenges of medical care. For a systematic review see \cite{medic2019evidence}. One of the tasks in which CDS can offer assistance is risk assessment.

There is a large number of works using machine learning methods on data extracted from Electronic Health Records (EHR) to predict future diagnoses, medication prescriptions and other medical events.
In \cite{miotto2016deep} authors used a three-layer stack of denoising autoencoders to derive a general-purpose patient representation from EHR data showing good prediction quality for severe diabetes, schizophrenia, and various cancers.
RNN are used on features extracted from EHR in \cite{choi2016doctor} to predict diagnosis and medication codes and in \cite{tomavsev2019clinically} to predict acute kidney injury.
In \cite{choi2017gram} authors used information extracted from EHR to build graph-based attention models for diagnosis prediction and heart failure prediction.

The COVID-19 pandemic led to development of multiple methods for risk assessment for patients admitted with COVID-19 using clinical features as input. The main disadvantage of this methods is the need for these features to be either imported directly from EHR or somehow extracted from textual data.
In \cite{abdulaal2020prognostic} authors use an artificial neural network for point-of-admission mortality risk scoring system from an array of possible prognostic factors.
Extreme Gradient Boosting (XGBoost) is used in \cite{vaid2020machine} to predict in-hospital mortality and critical events at different time windows on on continuous and one-hot encoded categorical features.
In \cite{haimovich2020development} authors used logistic regression to create  quick COVID-19 Severity Index scoring system using respiratory failure within 24 hours of admission as a target.
In \cite {liu2020covid} authors create a risk assessment model based on multiclass logistic regression using features extracted from patients’ demographic information, clinical symptoms, and contact history, blood tests and CT imaging results.

There are few works that use as input raw textual EHR data without feature extraction. In \cite{menger2018comparing} authors use RNN and CNN on clinical text available at the start of psychiatric admission to predict violence incidents. 
In \cite{obeid2020artificial} authors use CNN on telehealth system notes to predict the probability of COVID-19.

The current de-facto standard for most text-related tasks are self-attentive Transformer neural networks such as BERT. In \cite{devlin2018bert} authors show the effectiveness of such networks on Natural Language Processing (NLP) tasks especially after masked language modelling pretraining on unlabeled corpus of text. The drawback of this architecture is that the computational complexity of self-attention scales quadratically with the length of input sequences.

There are several works exploring the application of pretrained Transformer models to clinical texts, further increasing the prediction quality, but such models are still constrained by maximum input sequence length
In  \cite{alsentzer2019publicly} authors train BERT models on clinical texts with maximum sequence input length of 150 tokens.
In \cite{huang2019clinicalbert} authors train BERT model on clinical texts and predict hospital readmission. In this work the maximum input sequence length is 512 tokens, to process larger input sequences the authors split the full text into 512-sized chunks and process them separately, aggregating the results.

To overcome the computational problem of combining long input sequences and self-attentive models in \cite{beltagy2020longformer} authors present Longformer a modification of Transformer architecture using a combination of a global attention and a windowed local-context self-attention. Notably Longformer attention mechanism can act as a drop-in replacement for the self-attention mechanism in pretrained Transformers. This allows to use an existing pretrained BERT model and further fine-tune the model on much longer sequence length.

The aim of our proposed system is exactly the same as in \cite{marcos2020development} "to develop a machine-learning model to early identify patients who will die or require mechanical ventilation during hospitalization". The differences are in:
\begin{itemize}
\item data used: clinical and laboratory features obtained at admission \textit{vs} raw admission texts;
\item model: logistic regression, XBoost and random forests \textit{vs} Longformer;
\item study size: 1.3 thousand patients \textit{vs} 40 thousand;
\item validation metrics: ROC-AUC 0.85 \textit{vs} ROC-AUC 0.9+.
\end{itemize}

\section{The data}

For this work as input for the model we used anonymized textual data from EHR, available either at the time of admission (admission report) or during first 24 hours after admission (radiology reports, laboratory results, additional diagnoses). 
Admission reports differ in form and structure between medical institutions but generally contain all the relevant medical data known at admission time including: 
\begin{itemize}
\item initial complaints and anamnesis;
\item physical examination data;
\item laboratory and instrumental studies performed in the admission department;
\item routing list and treatment prescriptions;
\item any additional information at the discretion of the doctor in charge of admission
\end{itemize}

Other data available in EHR which medical specialists consider to be informative for the purposes of patient risk assessment include:
\begin{itemize}
\item  Radiology reports - Radiologist description of patient's chest CT, X-Ray or MRI.
\item  Lab reports - Results from laboratory tests including complete blood count with differential,  clinical urine analysis,  blood chemistry and others.
\item  Diagnoses - A list of diagnoses including secondary diagnoses such as hypertension, diabetes mellitus, asthma and others.
\end{itemize}

The common format for such data is a stream of events, each event corresponding to a single EHR form. A patients hospitalisation history can be processed as a sequence of forms, starting with admission report and ending either in discharge form or a death form.

For the purpose of this work we concatenated the admission report with all the additional text data available after 24 hours of hospital stay in chronological order. As a target we used the cases where the patient either died or was placed into artificial lung ventilation.

We used two datasets created from regional EHR systems. The first one referred hereafter as 'Dataset T' is from all public hospital of a region. The second one referred hereafter as 'Dataset K' is from a several major public hospital specialized in respiratory diseases located in another region.

\subsection{Dataset T}
Dataset T contains information about the hospitalization history of patients with pneumonia and was extracted from EHR data from a regional network of public clinic and hospitals. Full dataset consists of 103 million records from 2017 to 2019 each record corresponding to a single EHR form of a following structure: \textbf{<Patient ID>}, \textbf{<Clinic ID>}, \textbf{<Form name>}, \textbf{<Form text>}, \textbf{<ICD code>}. This data contains information both from primary care institutions and hospitals.

In addition to the baseline scenario starting with patient admitted with pneumonia and ending with either discharge or death we considered the following additional scenarios:
\begin{itemize}
\item  Patient admitted with other disease of respiratory system (ICD-10 codes J00-J99) but were subsequently diagnosed with pneumonia in the hospital. Such cases where included in the dataset.
\item Patient admitted to the hospital with suspicion of pneumonia but immediately discharged, usually after radiology report not confirming the suspicions. Such cases where excluded from dataset.
\item Patients admitted with pneumonia, discharged and readmitted again in a short period of time (less than a week) to continue treatment. Such split cases where considered as a single case.
\end{itemize}

The data from this dataset is noisy, some records are missing and the form names differ not only between institutions but also change in time within one institution. To extract the relevant data we made manual lists of names for relevant form types. The extracted form types  are described in Table~\ref{tab:forms}

\begin{table}
  \caption{EHR Form types}
  \centering
  \label{tab:forms}
  \begin{tabular}{lc}
    \toprule
    Form type & Contained information \\
    \midrule
    Admission & Initial patient information at admission time\\
    Rejection & Refusal of hospitalization\\
    Stay & Daily information about patient's status\\
    Radiology & Radiologist description of chest CT or MRI\\
    Discharge & Information about patient discharge\\
    Death& Information about patients death\\
  \bottomrule
\end{tabular}
\end{table}

We used the following procedure to extract the relevant cases from the stream of events imported from EHR:
\begin{enumerate}
\item For each patient select sequences of events which start with a disease of respiratory system diagnosis and have intervals of at most 10 days between sequential events.
\item If a patient sequence contains a discharge form and all the forms after this form do not contain pneumonia diagnoses, drop all the events after said discharge form.
\item If the event sequence contains rejection forms drop all the events before rejection form if they don't contain stay forms.
\item Drop the sequences that do not contain the pneumonia diagnoses.
\item Drop the sequences that do not contain the stay forms.
\end{enumerate}

The event sequences resulting from such procedure were considered well-formed cases if they contain an incoming form within 7 days from first event and contain a discharge form or a death report or mention of discharge or death in the last 4 days of the sequence. The cases that were not well-formed were either ongoing cases with patient still in the hospital or malformed cases with some forms missing.
We considered as positive targets the cases where the sequence contained a death form or a mention of ICU, artificial lung ventilation or death.

We split all the data for this dataset into cases treated in two biggest hospitals and cases treated in all the remaining hospitals. This groups are referred hereafter as 'Dataset T major' and 'Dataset T minor'. The statistics are presented in Table~\ref{tab:data_t}.

\begin{table}
  \caption{Dataset T (pneumonia cases) description }
  \centering
  \label{tab:data_t}
  \begin{tabular}{lcc}
    \toprule
    &Major & Minor\\
    \midrule
    Number of institutions & 2 & 13\\
    Number of admission forms & 11& 66\\
    Number of rejection forms & 6& 6\\
    Number of stay forms & 3 & 44\\
    Number of radiology forms & 11 & 15\\
    Number of discharge forms & 4 & 23\\
    Number of death forms & 5& 6\\
    \midrule
    Pneumonia cases formed & 5,311 & 9,613\\
    Pneumonia cases with radiology forms & 4,438 & 5,649\\
    Ongoing cases & 115 & 443\\
    Malformed cases & 170 & 1506\\
  \bottomrule
\end{tabular}
\end{table}

\subsection{Dataset K}
Dataset K was provided by several major public hospital specialized in respiratory diseases located in another region and contains information for the period of 2019-01-01 to 2020-11-01. This dataset contains only cases of patients diagnosed with COVID-19 or pneumonia. The data was presented in form \textbf{<Patient ID>}, \textbf{<Case ID>}, \textbf{<Hospitalization date>}, \textbf{<List of admission reports with dates>}, \textbf{<List of radiology reports with dates>}, \textbf{<List of lab reports with dates>}, \textbf{<List of ICD diagnoses with dates>}, \textbf{<ALV placement date>}, \textbf{<Death date>}, \textbf{<COVID-19 test result>}
This dataset did not require such extensive prepossessing as the regional one. We removed the cases without at least one admission report or with hospitalization date not set.

Due to a large number of false negative results for COVID-19 tests it was common practice in the hospitals providing the data to diagnose COVID-19 basing not only on test results but also on symptoms and chest radiography report. In total for 27 percent of patients diagnosed with COVID-19 the tests where negative. To explore whether the results of tests alone can be used to correctly select relevant training examples we used subsets of this dataset based either on COVID-19 test results (COVID-19 test) or the presence of COVID-19 diagnoses in patients list of diagnoses (COVID-19 diag).
We considered as positive targets the cases where either artificial lung ventilation date or death date was filled.

The statistics for the dataset are presented in Table~\ref{tab:data_k}. Number in parenthesis show the number of positive targets.

\begin{table}
  \caption{Dataset K pneumonia and COVID-19 cases}
  \centering
  \label{tab:data_k}
  \begin{tabular}{lccc}
    \toprule
    &2019 Jan-Dec& 2020 Jan-Jun & 2020 Jun-Nov\\
    \midrule
    Pneumonia & 13,501 (2,273)& 2,523 (218) & 814 (61)\\
    COVID-19 test & 0& 7,112 (1,120) & 2,238 (406)\\
    COVID-19 diag & 0 & 9,587 (1,312)& 3,901 (489)\\
  \bottomrule
\end{tabular}
\end{table}

\section{The method}
\subsection{Longformer}
Models with self-attention such as BERT are currently considered state of the art for multiple NLP tasks including text classification. Unfortunately the computational complexity of self-attention grows quadratically with input length so most models are pretrained with input lengths not exceeding 512 tokens. The length of admission forms routinely exceeds 512 tokens even without any additional data. The statistics for the datasets used in this study are presented in Table~\ref{tab:data_length}

\begin{table}
  \caption{Input sequence lengths}
  \centering
  \label{tab:data_length}
  \begin{tabular}{lcccccc}
    \toprule
    &&&\multicolumn{4}{c}{Percentiles} \\
    \multicolumn{1}{c}{Dataset} & Max & Mean & 0.25 & 0.5 & 0.75 & 0.99\\
    \midrule
    \multicolumn{7}{c}{Admission reports only}\\
    Regional 1 & 3,420 & 589 & 426 & 534 & 645 & 1,833 \\
    Regional 2 & 14,592 & 1,082 & 682 & 943 & 1,276 & 3,695 \\
    Hospital 2019 & 10,587 & 493 & 243 & 407 & 609 & 2,365 \\
    Hospital 2020 & 6,864 & 586 & 357 & 528 & 717 & 2,276 \\
    \midrule
    \multicolumn{7}{c}{Admission reports and additional data}\\
    Regional 1 & 3,763 & 808 & 649 & 767 & 888 & 2,114 \\
    Regional 2 & 14,750 & 1,527 & 1,074 & 1,340 & 1,725 & 4,681 \\
    Hospital 2019 & 10,587 & 898 & 445 & 771 & 1,150 & 3,263 \\
    Hospital 2020 & 6,868 & 945 & 559 & 841 & 1,197 & 2,897 \\
  \bottomrule
\end{tabular}
\end{table}

As shown in \cite{devlin2018bert} the performance of Transformer models is greatly improved by self-supervised pretraining on masked language modelling (MLM) task. For this work we used a BERT model already pretrained with MLM objective  on sequences from EHR texts with lengths up to 512. This model was converted to a Longformer architecture and pretrained on EHR sequences with lengths up to 8192 as described in next section. 
We used the implementation of Longformer from hugginface transformer library \cite{wolf2019huggingface}.
All models in this work where trained for 3 epochs with Adam optimizer with weight decay \cite{you2019large}, learning rate linearly decaying from 1e-5 to 0 and batch size 6.

\subsection{Pretraining Longformer}
Similarly to \cite{beltagy2020longformer}, to reduce computational costs we initialized Longformer from the basic BERT pretrained on EHR data. Besides, we did not initialized additional position embeddings randomly, but instead, we copied 512 positional embeddings of BERT 16 times.

We pretrained Longformer on MLM task with the mask probability equal to 0.15 on the corpus of patient histories which contained 170 thousand histories. These histories contained reports about the symptoms, syndromes, diseases, lab tests, and patient treatments. According to \cite{beltagy2020longformer} the model needs to learn the local context ﬁrst before starting to utilize the larger context. Thus, before training, we sorted our dataset by the number of tokens in history, so that the model first gets familiar with short patient histories and then adapts to longer histories. We used gradient checkpointing \cite{chen2016training} to fit a larger batch size into the memory of 32GB NVIDIA Tesla V100 GPU. We pretrained Longformer for 3 epochs with the batch size equal to 4 and the abovementioned default parameters. This pretraining process took 3 days on 1 GPU. 

\subsection{Baselines}
We used three BERT-based baselines to compare the effectiveness of Longformer model applied to whole input sequence vs BERT model applied to input sequence split in chunks of 512.
\begin{itemize}
    \item A BERT model applied to first 512 token chunk, effectively truncating the input (BERT).
    \item A BERT model applied to all 512 token chunks, the resulting contextualised embeddings vectors for first token of each chunk are averaged and passed to the linear classification layer (BERT-AVG)
    \item A BERT model applied to all 512 token chunks, the resulting contextualised embeddings vectors for first token of each chunk are used as input to Gated Recurrent Unit \cite{chung2014empirical} (GRU) RNN. The output of the GRU RNN after the last chunk is used as input for linear classification layer (BERT-RNN).
\end{itemize}

For the fairness of comparison we used the same pretrained Longformer used in this work for each of the baselines using maximum input sequence length of 512 instead of 8192. The resulting model is identical to BERT as it is using only global self-attention. Such setup allows us to fully use the benefits of masked language modelling pretraining for the baselines while allowing to estimate the effect of splitting the whole input sequence into smaller-sized chunks and applying self-attention on them separately.

\section{Experiments}
\subsection{Pneumonia}
We conducted experiments on risk assessment quality for model trained on pneumonia data, to see how well the proposed model generalizes across different hospitals in same region and between regions. Also we explored how the model trained on pneumonia cases generalizes for new disease with resembling symptoms such as COVID-19.
We used Dataset T major and cases from Dataset K dated 2019 as our train sets. For testing we used Dataset T major, Dataset T minor , and following subsets of Dataset K: all cases from 2019, all cases from 2020, cases from 2020 diagnosed with pneumonia, cases from 2020 diagnosed with COVID-19.
We trained two versions of the model for each training dataset, one only on admission data and another on all data available during first 24 hours after admission.
In this work we use the area under the ROC curve (ROC-AUC) performance measure.

The results are presented in Table~\ref{tab:pneumo_experiments} with first number in each cell showing the result for the model trained and tested on admission reports only, and the second number showing the result for the model trained with all data available during first 24 hours of hospital stay.

\begin{table*}
  \caption{ROC-AUC for pneumonia models trained and tested on admission reports only/admission reports with additional data}
  \centering
  \label{tab:pneumo_experiments}
  \begin{tabular}{c|cc|cccc}
    \toprule
    &\multicolumn{2}{c|}{\textbf{Dataset T}}&\multicolumn{4}{|c}{\textbf{Dataset K}} \\
    Train set/model &Major & Minor & 2019 & 2020 Pneumonia & 2020 COVID-19 & 2020 All\\
    \midrule
    Dataset T major&&&&&& \\
    \multicolumn{1}{r|}{Longformer}&-&91.7/\textbf{91.9}&\textbf{90.4}/88.3&90.9/90.1&88.7/87.8&89.2/88.3 \\
    \multicolumn{1}{r|}{BERT}&-&86.7/-&88/-&89.2/-&86.7/-&87.2/- \\
    \multicolumn{1}{r|}{BERT-AVG}&-&90.4/-&88,6/-&88,5/-&87.8/-&88.1/- \\

    \multicolumn{1}{r|}{BERT+RNN}&-&90.6/-&87.5/-&89.5/-&87.2/-&87.7/- \\
    \midrule
    Dataset K 2019 &&&&&& \\
    \multicolumn{1}{r|}{Longformer}&85.6/85.0&86.2/84.2&-&\textbf{91.3/91.3}&\textbf{90.1}/89.7&\textbf{90.4}/90.1 \\
    \bottomrule
  \end{tabular}
\end{table*}

Our experiments show that the Longformer architecture outperforms all baselines on the risk prediction task by a significant margin. As expected truncated BERT is worse than BERT-AVG and BERT-RNN due to information loss during truncation of input sequence to 512 tokens.

Unsurprisingly the model trained on more similar data provides for better result. The best ROC-AUC score for Dataset T minor is achieved by a model trained on Dataset T major. Best result for Dataset K 2020 is achieved by a model trained on Dataset K 2019. 

Interestingly the results for model trained on Dataset T major are very similar to the result of model trained on Dataset K 2019 when testing on Dataset K 2020. The reverse is not true and there is a noticeable quality drop when using model trained on Dataset K 2019 to predict cases from Dataset T. The reason of this difference may be in that the Dataset T contains more diverse form variants, so the model trained on this data has greater generalization capability when presented with data from another institutions.

The models trained on pneumonia data perform slightly worse when presented with cases of COVID-19 within the same dataset. The performance drops significantly when the change in disease and change in dataset occurs simultaneously. Our experiments also show that for the task of risk assessment the additional data received during first 24 hours of hospital stay does not increase the quality of predictions for models trained on pneumonia data only.

All models benefit from adding data available during first 24 hours of hospital stay with the exception of the model trained on data of patients with positive COVID-19 test and tested on patients without COVID-19. This shows that the model successfully extracts relevant medical data both from admission reports and from additional textual data.

\subsection{COVID-19}

To determine what training strategy would be most effective in cases of COVID-19 and pneumonia we trained models on three subsets of Dataset K:
\begin{itemize}
    \item All patients with positive COVID-19 test for period from January to June 2020
    \item All patients with COVID-19 diagnosis for period from January to June 2020
    \item All patients for period from January 2019 to June 2020
\end{itemize}
We tested the models on the following subsets of Dataset K:
\begin{itemize}
    \item All patients with positive COVID-19 test for period from July to November 2020
    \item All patients with COVID-19 diagnosis for period from July to November 2020
    \item All patients with Pneumonia diagnosis for period from July to November 2020
    \item All patients for period from July to November 2020
\end{itemize}
As with pneumonia models we trained two versions for each model. First one on admission reports only and the second on all data available during 24 hours after admission. The results are shown in Table~\ref{tab:covid_experiments} 

\begin{table*}
  \caption{ROC-AUC for Dataset K  July 2021 - November 2021}
  \centering
  \label{tab:covid_experiments}
  \begin{tabular}{c|cccc}
    \toprule
    Train set & COVID-19 test & COVID-19 diag & Pneumonia & All \\
    \midrule
    COVID-19 test 2020 Jan-June&88.7/88.7&90.1/90.1&87.5/85.3&90.0/89.8 \\
    COVID-19 diag 2020 Jan-June&88.6/\textbf{90.0}&90.1/\textbf{91.2}&85.5/85.8&89.8/90.6 \\
    All data 2019 Jan- 2020 June&89.3/89.5&90.6/90.7&87.3/\textbf{89.6}&90.4/\textbf{90.7} \\
    \bottomrule
  \end{tabular}
\end{table*}

Using training dataset of patients with positive COVID-19 test leads to generally poor results across all test sets with additional information not providing any performance boost.

For patients with positive COVID-19 test or COVID-19 diagnosis the best results are achieved using the model trained on data from patients diagnosed with COVID-19. For pneumonia patients and combined pneumonia and COVID-19 patients the best results are achieved when training on all data available up to June 2020. 
Additional data from first 24 hours after hospitalisation improves risk assessment on all testing datasets for both models trained on COVID-19 diagnosed patients and on all patients. 

\section{Discussion}

COVID-19 has identified many problems in healthcare systems around the world. Almost all hospitals were faced with difficulties in the management of resources and staff. Doctors were forced to treat record number of patients in harshest conditions. Specialists who do not work with respiratory diseases in their daily practice have often been drafted to treat patients with pneumonia and COVID-19. All these factors created the preconditions for sub-optimal routing of patients in the hospital. In this situation patients at low risk of complications could be misdirected to intensive care units while patients at high risk could be left without due attention.

The goal of our work was to develop a system that could help doctors in assessing the risks of complications in patients with COVID-19 and pneumonia. This system should be easy to integrate into hospital electronic medical record systems. As well it should be able to deal with the high variability in forms of medical examinations and diagnostic tests.

To achieve this goal, we applied the Longformer model to raw admission forms and other textual information available during first 24 hours of hospital stay. We were able to achieve high prediction quality in the task of predicting the risk of death or the need for artificial ventilation of lungs in patients hospitalized with pneumonia or COVID-19. The efficiency of this model has been shown on retrospective datasets. 

We showed that the risk assessment can be built on the unstructured texts of electronic medical records, without any additional preprocessing and feature extraction. This separates our system from other prognostic systems \cite{abdulaal2020prognostic, haimovich2020development, liu2020covid, vaid2020machine, marcos2020development}, which make predictions on previously identified features. Use of the unstructured text allows  to easily integrate our system into clinical practice regardless of EHR system used by the hospital. 

Our model is resilient to the variability of forms of medical examinations, which allows it to be used in other hospitals without retraining the model. However the optimal results are achieved when training the model on the data gathered in the same hospital and same disease.

We have demonstrated that using a Longformer architecture that combines global and local attention will improve performance compared to BERT architectures when applied to long medical texts. Longformer combination of global and local self-attention allows all available patient data to be used simultaneously without splitting it into separate blocks. We believe that the more accurate prediction of this architecture compared to BERT baselines is achieved using by using self-attention on the broad context of the complete input sequence. 

However, along with the obvious advantages, one of the problems of the proposed model is the absence of interpretability. Therefore, hospitals must weigh the risks and benefits before implementing our model into their workflows.


Our Longformer-based deep learning model assesses with high accuracy the risks of artifical ventilation of  lungs or death in patients hospitalized with pneumonia or COVID-19. As our model uses only raw text as input it can be quickly integrated with any electronic health records system. This property is critical for many hospitals in the world particularly in developing countries which still store their data as raw texts only. 

\bibliographystyle{ACM-Reference-Format}
\bibliography{references}

\end{document}